\documentclass{article} 
\usepackage{iclr2024_conference,times}

\usepackage{amsmath,amsfonts,bm}

\def\eqref#1{equation~\ref{#1}}

\def\1{\bm{1}}

\DeclareMathAlphabet{\mathsfit}{\encodingdefault}{\sfdefault}{m}{sl}
\SetMathAlphabet{\mathsfit}{bold}{\encodingdefault}{\sfdefault}{bx}{n}

\usepackage{hyperref}
\usepackage{url}
\usepackage{inconsolata}
\usepackage{CJKutf8}
\usepackage{graphicx}
\usepackage{amssymb}
\usepackage{multirow}
\usepackage{pifont}
\usepackage{tabu}
\usepackage{booktabs}
\usepackage{color}
\usepackage{subcaption}
\usepackage{tablefootnote}
\usepackage{wrapfig}
\usepackage{caption}
\usepackage{threeparttable}
\usepackage{colortbl}
\title{SkyMath: Technical Report}

\author{Liu Yang, Haihua Yang, Wenjun Cheng, Lei Lin, Chenxia Li, Yifu Chen, Lunan Liu, \\
\textbf{Jianfei Pan, Tianwen Wei, Biye Li, Liang Zhao, Lijie Wang, Bo Zhu, Guoliang Li, } \\
\textbf{Xuejie Wu, Xilin Luo, Rui Hu$^{\dagger}$} \\
\\ \textbf{Kunlun Inc.} \\
\texttt{rui.hu@kunlun-inc.com} \\
}

\iclrfinalcopy 
\begin{document}

\maketitle

\begin{abstract}
Large language models (LLMs) have shown great potential to solve varieties of natural language processing (NLP) tasks, including mathematical reasoning. In this work, we present SkyMath, a large language model for mathematics with 13 billion parameters. By applying self-compare fine-tuning, we have enhanced mathematical reasoning abilities of Skywork-13B-Base remarkably. On GSM8K, SkyMath outperforms all known open-source models of similar size and has established a new SOTA performance. On dataset MATH and out-of-domain dataset CMath, SkyMath also achieves a high accuracy rate, showing remarkable generalizability to varieties of math problems.
\end{abstract}

\section{Introduction}
Nowadays, Large language models (LLMS) are increasingly being applied to all kinds of complex tasks, including content creation \citep{gardner2023llark,abs-2309-15112,abs-2304-10592,abs-2305-06500,abs-2304-08485}, code generation \citep{NijkampPHTWZSX23,abs-2305-07922,FriedAL0WSZYZL23,abs-2107-03374,abs-2305-06161,zheng2023codegeex}, multi-turn conversations \citep{abs-2201-08239,llama2,bai2023qwen,abs-2309-10305,zeng2023glm-130b,abs-2308-15930}, mathematical reasoning \citep{azerbayev2023llemma,abs-2309-05653,abs-2309-12284,abs-2308-09583,QiaoO0CYDTHC23,abel,fu2023kwaiyiimath}, and knowledge-based question answering \citep{abs-2306-16092,abs-2303-17564,abs-2306-06031,abs-2304-06975,zhu2023ChatMed}, and they have the potential to revolutionize the fields of natural language processing and natural language understanding \citep{openai2022chatgpt,openai2023gpt4}. Moreover, compared to traditional AI methods, LLMs gain unparalleled advantages in these landscapes. Generative Artificial Intelligence (GenAI) fueled by LLMs is just around the corner.

Despite their impressive capabilities, LLMs come with a series of challenges and issues. Mathematical reasoning is one of subfields worth exploring. In the context of evaluating the reasoning capabilities of LLMs, complex mathematical reasoning serves as a crucial benchmark. Moreover, mathematical reasoning tasks are considered to be one of the major gaps between closed-source models like ChatGPT \citep{openai2022chatgpt} or GPT4 \citep{openai2023gpt4} and open-source models like LLaMA \citep{abs-2302-13971,llama2}. To achieve advanced mathematical reasoning capabilities, numerous open-source models have made attempts: Kuaishou introduces the KwaiYiiMath by applying Supervised Fine-Tuning (SFT) and Reinforced Learning from Human Feedback (RLHF) and constructed a small-scale Chinese primary school mathematics test set (named KMath) \citep{fu2023kwaiyiimath}; Microsoft presentes wizardMath by applying their proposed Reinforcement Learning from Evol-Instruct Feedback (RLEIF) method to the domain of math \citep{abs-2308-09583}; Xiang et al. introduce MAmmoTH by training the model on their self-developed dataset MathInstruct \citep{abs-2309-05653}; Zhangir et al. presente LLEMMA by continue pretraining Code Llama on Proof-Pile-2, a mixture of scientific papers, web data containing mathematics, and mathematical code \citep{azerbayev2023llemma}; Yu et al. \citep{abs-2309-12284} proposes MetaMath trained on a new dataset called MetaMathQA which is build by bootstrapping mathematical questions by rewriting the question from multiple perspectives. Despite the great success, most existing open-source LLMs are still far away from satisfactory for solving mathematical problems due to the complex reasoning procedures, and preform poorly on GSM8K and MATH when compared to open-sourced models.

To bridge this gap, we propose SkyMath, a finetuned language model that specializes in mathematical reasoning, by applying our improved data augmentation techniques and SFT process to the finetuning of Skywork-13B-Base. The main process consists of three parts, as follows: 1. Finetuning the Skywork-13B-Base model on open-source datasets; 2. Constructing a new dataset for math by improved data augmentation techniques; 3. Reconstructing math dataset by applying our proposed self-check techniques. Experimental results show that SkyMath outperforms many open-source models in similar sizes on two mathematical benchmarks, namly GSM8k \citep{abs-2110-14168} and MATH \citep{HendrycksBKABTS21}.

The paper is structured as follows. Section 2 provides an overview of related work including LLMs and LLMs for mathematics. Section 3 introduces the methodology of Skywork-13B-Math including sample construction methods and process of supervised fine-tuning . Then, the experimental validations and comparisons are made in Section 4 and Section 5. The conclusion is drawn in Section 6.

\section{Related Work}

\paragraph{Large Language Models}
Recently, LLMs have achieved substantial progress in natural language processing tasks, benefiting from high-quality, diverse textual data, and enormous model parameter counts. Researchers have found that LLMs perform better on downstream tasks \citep{scaling}. 
LLMs, which pretrained on extensive textual data and fine-tuned for specific tasks, demonstrate remarkable capabilities in numerous natural language processing tasks compared to smaller models \citep{DevlinCLT19,WeiTBRZBYBZMCHVLDF22}. The significant capabilities of large models mainly lie in three aspects: 1. In-context learning. 2. Instruction following. 3. Step-by-step reasoning \citep{Survey}. The GPT-3 \citep{BrownMRSKDNSSAA20}, a language model that has tens of billions parameter counts, exhibits significant performance improvements in few-shot, one-shot, and zero-shot learning tasks through in-context learning. However, the effectiveness of in-context learning capability depends on different downstream tasks. The open-source model Llama2 \citep{llama2}, with a pre-trained context length of up to 4k, demonstrates robust contextual understanding in tasks such as summarization, multi-turn dialogues, and reading comprehension. LLMs handle different downstream tasks efficiently, but sometimes these pre-trained models still struggle to comprehend human instructions \citep{Ouyang0JAWMZASR22}. It is highly challenging for LLMs to understand and follow specific instructions in the complex natural language tasks. To overcome the challenge, Instruction Tuning (IT) \citep{abs-2308-10792} and Chain of Thought (CoT) \citep{Wei0SBIXCLZ22} have been proposed. Specifically, IT aims to utilize the constructed pairs whose formats are [instruction, output] to fine-tune the pre-trained language model. Moreover, the adoption of multitask instruction blending in fine-tuning has demonstrated the effectiveness of IT techniques in unseen tasks \citep{SanhWRBSACSRDBX22,Ouyang0JAWMZASR22,WeiBZGYLDDL22}. Numerous studies show that high-quality, diverse instructions can effectively improve the performance of LLMs in natural language tasks \citep{WangKMLSKH23,ZhouMHPPCB23,WeiBZGYLDDL22,abs-2307-06290}. CoT is a method that aims to improve the performance of LLMs by providing detailed reasoning processes as training inputs. By enabling the model to follow a step-by-step process, CoT helps LLMs to break down complex problems into smaller ones and accumulate small victories to achieve greater success \citep{Wei0SBIXCLZ22}. This approach has led to significant improvements in the reasoning capabilities of LLMs, especially in mathematical reasoning and decision-making tasks. The success of CoT and IT technologies in Self-Supervised Fine-Tuning (SFT) has opened up new possibilities for enhancing the performance of LLMs. Numerous models, including InstructonGPT \citep{Ouyang0JAWMZASR22}, BLOOMZ \citep{MuennighoffWSRB23}, WizardLM \citep{abs-2304-12244}, ChatGLM2 \citep{DuQLDQY022}, and Vicuna \citep{vicuna2023}, have acquired powerful reasoning abilities, allowing them to perform complex tasks with remarkable accuracy. In addition, by training LLMs on data specific to a particular domain, such as medical or legal, it is possible to obtain models that excel in that domain. Models obtained in this way include InstructDial \citep{GuptaJYMEB22}, Radiology-GPT \citep{abs-2306-08666}, Goat \citep{abs-2305-14201}, WizardCoder \citep{abs-2306-08568}.

\paragraph{Large Language Models for Mathematical reasoning}
Mathematical reasoning is one of the essential abilities required for LLMs. To enhance mathematical reasoning capabilities of base LLMs, a group of researchers conducted a comprehensive study. Based on CoT, a series of works are carried out to optimize the reasoning paths \citep{0002WSLCNCZ23,FuPSCK23,abs-2210-11610}. Wang et al. introduces Self-Consistency, employing the results of multiple reasoning paths to enhance the accuracy of inference through consistent answers \citep{0002WSLCNCZ23}. Fu et al. proposes Complexity-based CoT, achieving improved performance in mathematical reasoning tasks through a CoT involving multiple complex reasoning steps \citep{FuPSCK23}. In terms of prompting engineering, Zheng et al. combines CoT to generate answers, utilizing the answer from the previous step as a prompt for the next step, gradually guiding LLMs to generate correct answers \citep{abs-2304-09797}. Luo et al. proposes reinforced evol-instruct, a method that combines evol-instruct with reinforcement learning to enhance the reasoning performance of LLMs \citep{abs-2308-09583}. Madaan et al. introduces Self-Refine, which a single LLM serves simultaneously as a generator, refiner, and feedback provider, iteratively refining itself to enhance its capabilities in mathematical reasoning \citep{abs-2303-17651}. Yu et al. constructs a new dataset, MetaMathQA, through data augmentation of rewriting of mathematical problems from multiple perspectives \citep{abs-2309-12284}. Fu et al. enhances the mathematical reasoning capabilities of LLMs through a combination of supervised fine-tuning and reinforcement learning with human feedback (RLHF) \citep{fu2023kwaiyiimath}. Based on Cot and Program of Thoughts (PoT) \citep{abs-2211-12588}, Yue et al. constructs a new dataset namly MathInstruct and uses it in SFT. PoT utilizes an external interpreter (e.g., Python interpreter) to compute answers for complex mathematical problems \citep{abs-2309-05653}. Azerbayev et al. continues pretraining on Proof-Pile-2, a dataset containing mathematical web data and mathematical code, and uses PoT to obtain an accurate answer \citep{azerbayev2023llemma}.
\section{Method}
\begin{figure*}[h]
 \centering
 \includegraphics[width=1.0\linewidth]{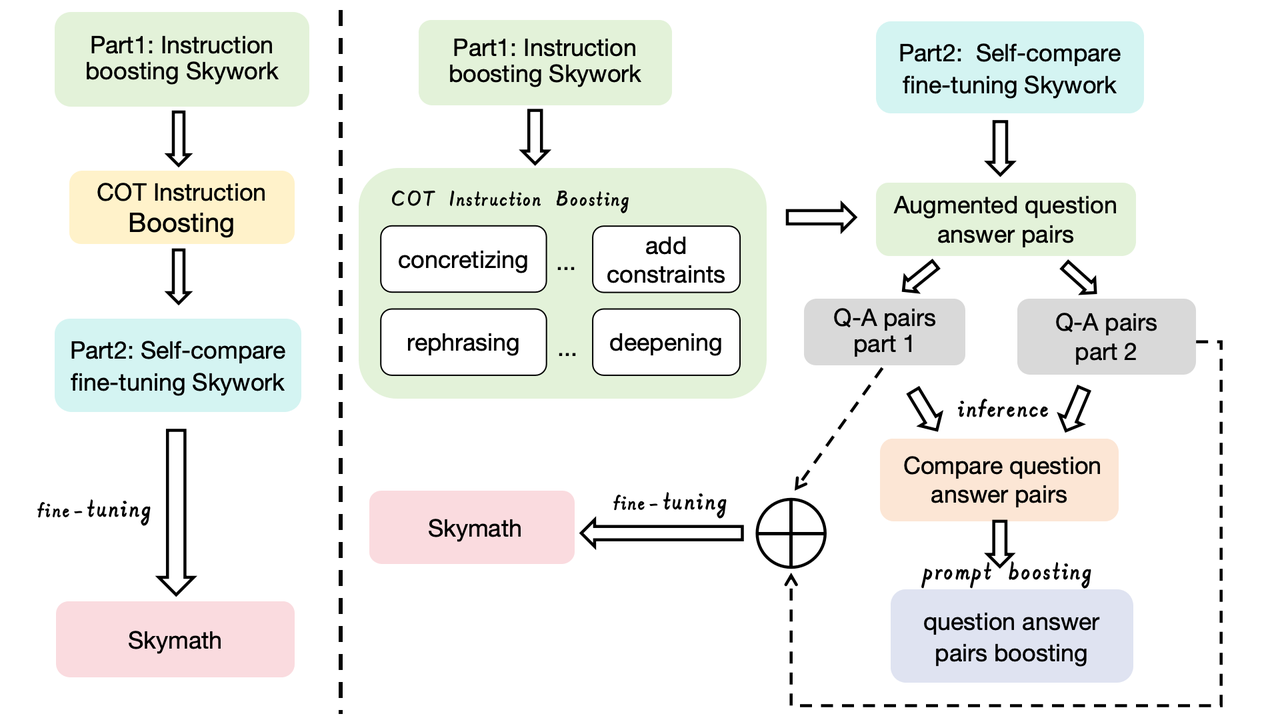}
 \caption{The Overview Architecture of SkyMath}
 \label{fig:framework}
\end{figure*}

In this section, we introduce SkyMath in detail. As shown in Figure~\ref{fig:framework}, our method mainly contains two steps:\\
1. Instruction boosting.\\
2. Self-compare fine-tuning.\\
\subsection{Instruction Boosting}
Previous work has shown math instructions with various complexities could make remarkable effect on math LLMs training \citep{abs-2308-09583}. Therefore, the first step we do is constructing a dataset of both high quality and diversity.\\
1. We first collect a math dataset from different sources, including different levels, in both Chinese and English. \\
2. Then inspired by WizardLM \citep{abs-2304-12244} and MetaMath \citep{abs-2309-12284}, we adapt instruction boosting, namely 1) concretizing, 2) adding constraints, 3) deepening and 4) rephrasing, to the question augmentation process.\\
3. We use LLMs to generate responses for augmented questions.\\
4. Correctness check.\\
By now, we have got a dataset of high complexity.

\subsection{Self-compare Fine-tuning}
Progressive-Hint Prompting (PHP) \citep{abs-2304-09797} enables multiple interactions between users and LLMs by using previously generated answers as hints to progressively guide LLMs toward the correct answers. Inspired by this, we believe introducing previous answers to the training process also has an effect. We hope the model can compare its previous answers with ground truth and correct its specific errors through training. 
\begin{figure*}[h]
 \centering
 \includegraphics[width=1.0\linewidth]{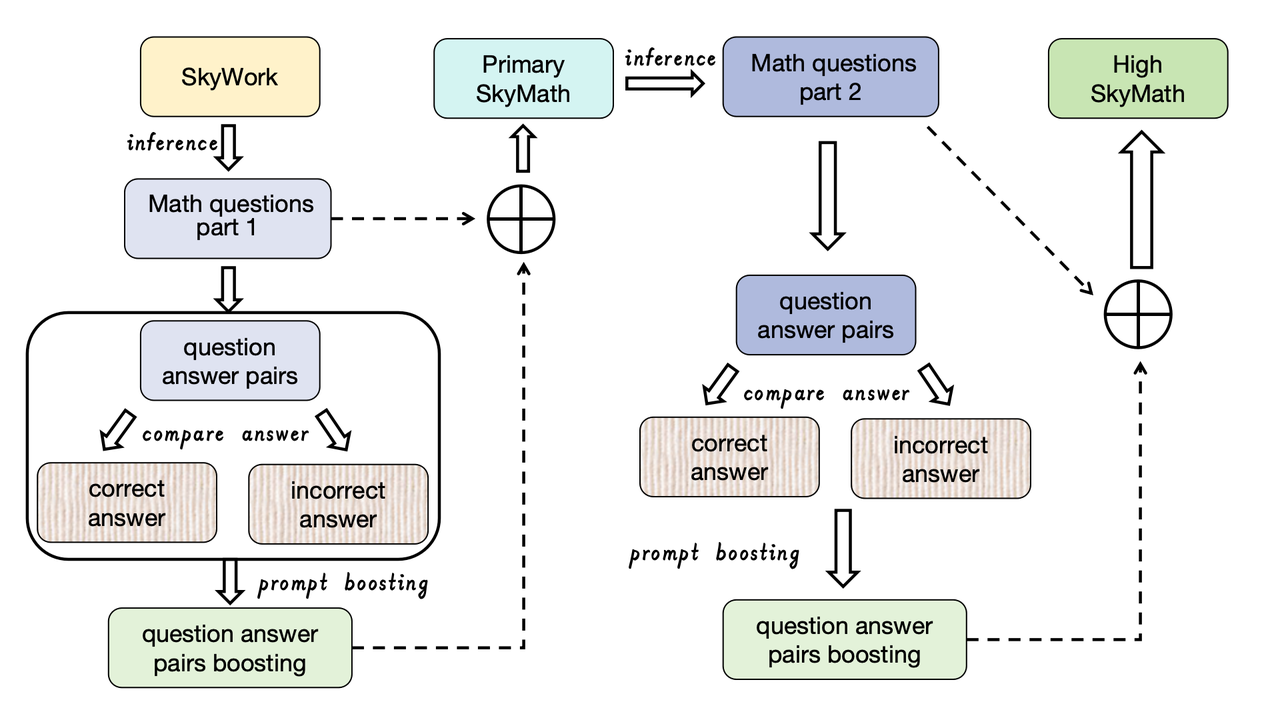}
 \caption{Self-compare Fine-tuning}
 \label{fig:selfcompare}
\end{figure*}

\begin{figure*}[h]
 \centering
 \includegraphics[width=1.0\linewidth]{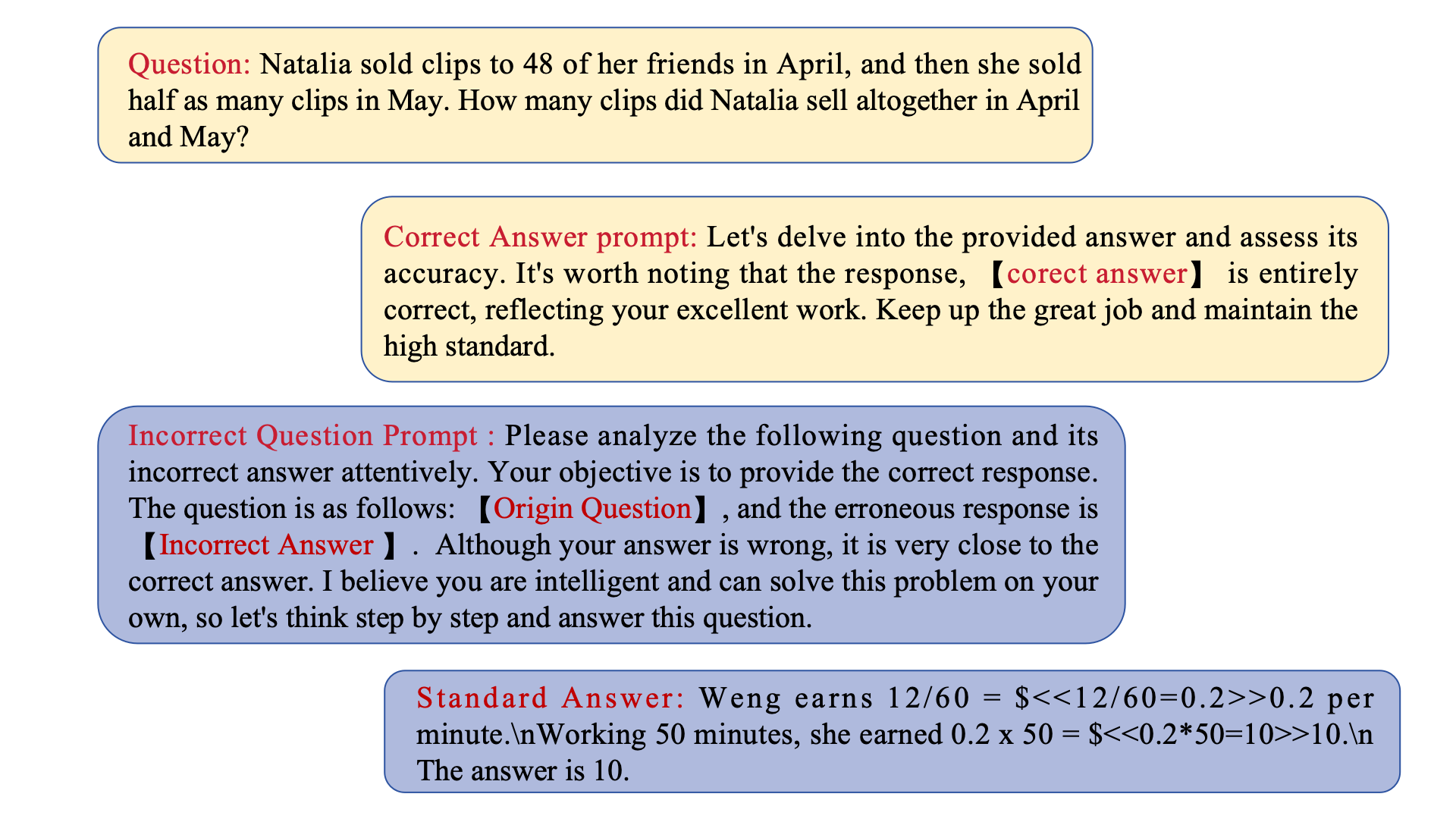}
 \caption{Self-compare Prompts}
 \label{fig:self-compare-prompts}
\end{figure*}

As shown in Figure~\ref{fig:selfcompare}, self-compare fine-tuning contains four steps:\\
1. For each question, ask the LLM to give an answer.\\
2. 2. Construct self-compare prompts, as shown in Figure~\ref{fig:self-compare-prompts}.\\
3. Combine the data with the origin dataset.\\
4. Fine-tuning.\\
Like human beings, we believe LLMs tend to make different mistakes as their abilities improve. Therefore, in practice we divide the origin dataset into several sub-datasets thus we can repeat self-compare fine-tuning more than once. 

\section{Experiments}

\subsection{Evaluation Benchmarks}
We mainly evaluate SkyMath on two popular mathematical reasoning benchmarks: GSM8k \citep{abs-2110-14168} and MATH \citep{HendrycksBKABTS21}. In order to see the performance of SkyMath on the out-of-domain dataset and in chinese, we also evaluate our model on CMath \citep{abs-2306-16636}, for completeness.

GSM8k contains 7,473 training data and 1,319 test data, mainly on grade school level English math problems. Each problem takes between 2 and 8 steps to solve, and solutions primarily involve performing a sequence of elementary calculations using basic arithmetic operations. 

MATH is much more challenging. It contains 7,500 training data and 5,000 test data, spans seven subjects including Prealgebra, Algebra, Number Theory, Counting and Probability, Geometry, Intermediate Algebra, and Precalculus.

CMath is a Chinese Elementary School Math Word Problems (CMATH) dataset, which contains 1.7k elementary school-level math word problems with detailed annotations, sourced from actual Chinese workbooks and exams. We use it as an out-of-domain dataset.

\subsection{Model and Baselines}
We use SkyWork-13B as base model and correspondingly choose models of the same size and has already been open source for comparison. Therefore we choose LLaMA1 \citep{abs-2302-13971}, LLaMA2 \citep{abs-2302-13971}, BaiChuan1 \citep{abs-2309-10305}, BaiChuan2 \citep{abs-2309-10305}, WizardMath \citep{abs-2308-09583}, GAIRMath-Abel \citep{abel}, and MetaMath \citep{abs-2309-12284} as baselines.

\section{Main Results}
\begin{table}[h]
\centering
\caption{Results of pass@1 (\%) on GSM8k, MATH and CMath}
\label{table:performance}
\begin{tabular}{lcccc}
\toprule
Model         & \#Params & GSM8K & Math  & CMath \\
\midrule
LLaMA1 \citep{abs-2302-13971}       & 13B      & 17.8  & 3.9   & -     \\
LLaMA2 \citep{abs-2302-13971}       & 13B      & 28.7  & 3.9   & -     \\
BaiChuan1 \citep{abs-2309-10305}    & 13B      & 26.76 & 4.84  & 51.33 \\
BaiChuan2 \citep{abs-2309-10305}    & 13B      & 52.77 & 10.08 & -     \\
WizardMath \citep{abs-2308-09583}   & 13B      & 63.9  & 14.0  & 50.83 \\
GAIRMath-Abel \citep{abel} & 13B      & 66.41 & 17.34 & -     \\
MetaMath \citep{abs-2309-12284}     & 13B      & 72.3  & 22.4  & -     \\
\textbf{SkyMath}       & 13B      & \textbf{72.33} & \textbf{16.98} & \textbf{77.27} \\
\midrule
\end{tabular}
\end{table}

Evaluating results are shown in Table~\ref{table:performance}. SkyMath outperforms all baselines on GSM8K, thus we have established a new SOTA performance across open-source LLMs of similar size. On the MATH dataset, which is highly challenging, SkyMath also achieves a high accuracy rate. Meanwhile, on the out-of-domain dataset CMath, SkyMath's performance shows remarkable generalizability to both unseen math problems and Chinese math problems.
\section{Conclusion and Future Work}
This paper introduces SkyMath, a mathematics model fine-tuned with self-compare. Evaluating results show SkyMath achieves SOTA performance across all existing open-source LLMs of similar size on GSM8K and shows remarkable generalizability to out-of-domain math problems. Considering we didn't use tools or reward models, the result is even more difficult.\\
\textbf{Future Work.} Although our model achieves impressive performance on GSM8K, we still fall far behind models like GPT-4. In the future, we will explore more methods to improve the abilities of our model.

\bibliography{custom}
\bibliographystyle{iclr2024_conference}


\end{document}